\documentclass[10pt,twocolumn,letterpaper]{article}

\usepackage[pagenumbers]{cvpr}

\usepackage{cuted}
\usepackage{capt-of}

\definecolor{cvprblue}{rgb}{0.21,0.49,0.74}
\usepackage[
    breaklinks,
    colorlinks,
    allcolors=cvprblue
]{hyperref}

\title{
OptimusMesh: Compact Autoregressive Mesh Generation from Point Clouds via
Sparse Latent Pivots
}

\author{
Mazhar Iqbal$^{1}$ \quad
Xuanmeng Sha$^{1}$ \quad
Naoya Chiba$^{1}$ \quad
Yuki Uranishi$^{1}$ \quad
Tomohiro Mashita$^{2}$\\[0.4em]
$^{1}$The University of Osaka \qquad
$^{2}$Osaka Electro-Communication University\\
{\tt\small
mazharicp786@gmail.com,\;
shaxuanmeng@gmail.com,\;
chiba@nchiba.net}\\
{\tt\small
yuki.uranishi.cmc@osaka-u.ac.jp,\;
mashita@osakac.ac.jp}
}

\begin{document}

\maketitle

%%%%%%%%% TEASER FIGURE
\begin{strip}
\centering
\vspace{-3mm}

\includegraphics[
    width=5.08\textwidth,
    height=71mm,
    keepaspectratio
]{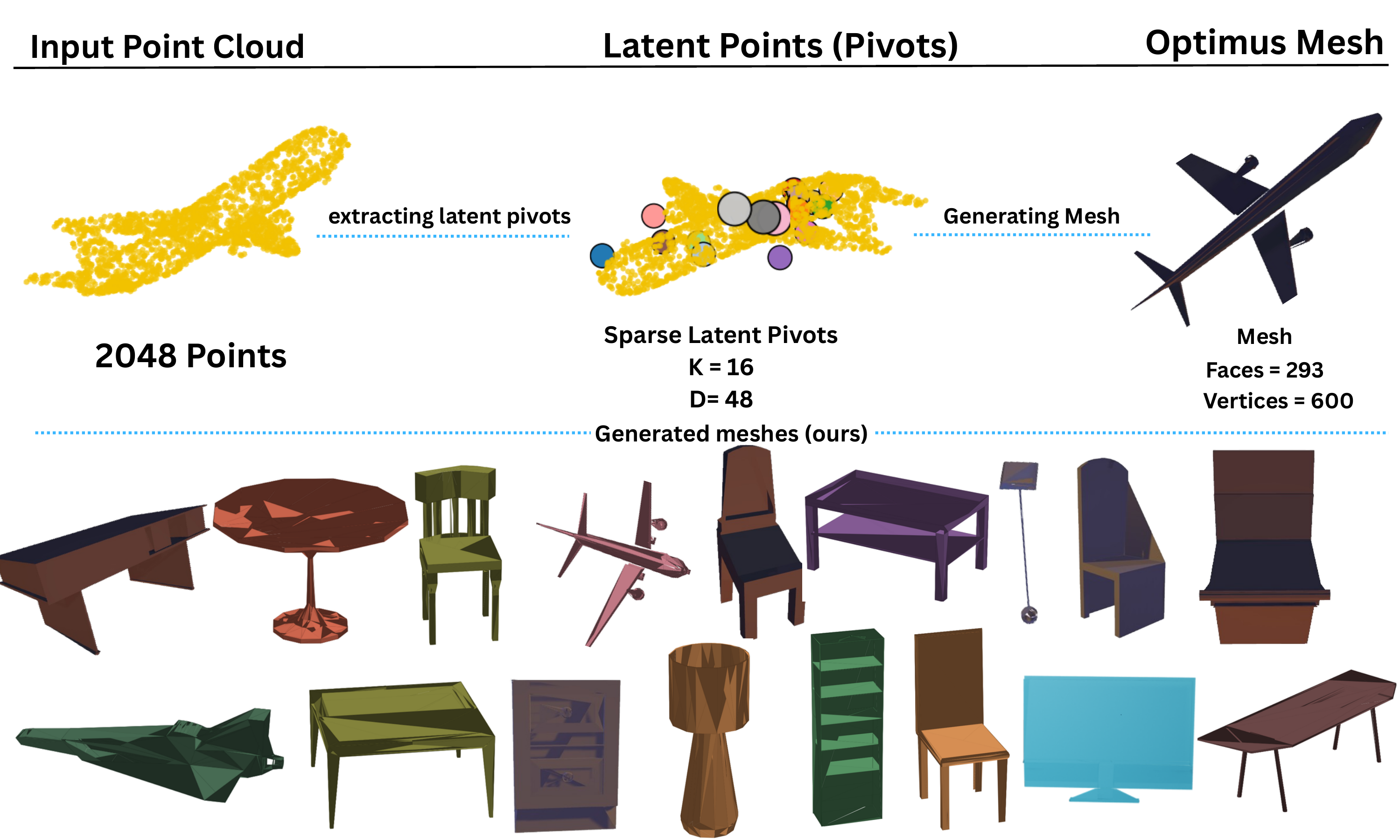}

\vspace{-2mm}

\captionof{figure}{
Starting from an input point cloud of 2,048 oriented points,
\textbf{OptimusMesh} extracts 16 sparse latent pivots, each consisting of a
3D position and 45 learned feature channels, and directly generates a compact
triangle mesh by autoregressively predicting vertices followed by triangular
faces. Additional meshes generated by our method are shown below.
}
\label{fig:teaser}

\vspace{-3mm}
\end{strip}

%%%%%%%%% MAIN PAPER
\begin{abstract}
Generating compact and geometrically faithful 3D meshes directly from point
clouds remains a fundamental challenge. Point clouds are unordered and sparse,
whereas meshes exhibit irregular structure and varying topology. As a result,
many existing approaches rely on implicit representations followed by surface
extraction or reconstruction. Although effective, these pipelines can produce
dense or over-smoothed meshes, often requiring computationally expensive
post-processing and simplification. We present \textbf{OptimusMesh}, a
framework for direct compact triangle mesh generation from point clouds using
sparse latent pivot conditioning. Our key idea is to compress $2{,}048$
oriented input points into only $16$ sparse latent pivots, reducing the
geometric conditioning set by $128\times$. These pivots provide a compact
structural representation shared across a two-stage autoregressive framework
that first generates mesh vertices and then predicts triangular faces
conditioned on the generated vertices and the same pivots. Compared with the
evaluated recent point-cloud-conditioned autoregressive methods, which use
$257$ decoder-conditioning tokens, OptimusMesh uses only $16$, yielding a
$16.1\times$ shorter conditioning sequence. Experiments show that OptimusMesh
produces the most compact outputs among the compared recent autoregressive
methods, using $25.7\%$--$94.1\%$ fewer faces while maintaining competitive
geometric fidelity and distributional quality. Project page: \url{https://mazharcyber.github.io/optimusmesh-project-page/}
\end{abstract}
\section{Introduction}
\label{sec:intro}

The field of 3D mesh generation has witnessed remarkable advancements in
recent years in 3D computer vision and computer graphics
~\cite{Poole2023DreamFusion,Hong2024LRM,Xu2024InstantMesh}. 3D meshes are a
fundamental representation of 3D geometry and are extensively used in video
games, movies, augmented and virtual reality, and architectural modeling. In
many practical settings, meshes must remain compact to meet memory, bandwidth,
and real-time rendering constraints, while also supporting efficient
level-of-detail (LOD) rendering and transmission.

Meshes represent object geometry through vertices, edges, and faces, but their
irregular connectivity introduces intrinsic challenges for generative models.
Mesh topology can vary widely across object instances, even within the same
category. Consequently, template-based approaches
~\cite{Gupta2020NeuralMeshFlow,Jakab2021KeypointDeformer,
Liu2021DeepMetaHandles,Wang2018Pixel2Mesh,Wen2019Pixel2MeshPlusPlus,
Yifan2020NeuralCages} that deform meshes with predefined connectivity can be
limited in representing diverse shapes and large topological variations.

Many recent methods rely on neural 3D representations such as
triplanes~\cite{Hong2024LRM,Li2024Instant3D} and
FlexiCubes~\cite{Xu2024InstantMesh}, followed by surface extraction. Although
effective, the resulting meshes can be dense or over-smoothed and may require
additional simplification. Autoregressive methods instead generate explicit
mesh structure directly. PolyGen~\cite{Nash2020PolyGen},
MeshGPT~\cite{Siddiqui2024MeshGPT}, MeshAnything~\cite{Chen2025MeshAnything},
MeshAnythingV2~\cite{Chen2025MeshAnythingV2},
FastMesh~\cite{Kim2026FastMesh}, DeepMesh~\cite{Zhao2025DeepMesh}, and
MeshRipple~\cite{Lin2026MeshRipple} improve mesh tokenization, topology modeling,
or output sequence efficiency. For point-cloud inputs, reconstruction methods
such as Screened PSR~\cite{Kazhdan2013ScreenedPoisson},
SAP~\cite{Peng2021ShapeAsPoints}, and
Point2Mesh~\cite{Hanocka2020Point2Mesh} can recover geometry, but often rely on
dense extraction or costly per-shape processing. In contrast, the efficiency
of the geometric conditioning supplied to direct mesh generators remains less
explored.

In this work, we propose \textbf{OptimusMesh}, a framework for direct compact
mesh generation from point clouds using sparse latent pivot conditioning. A
point-cloud encoder~\cite{Lyu2023SLIDE} compresses $2{,}048$ oriented input
points into only $K{=}16$ spatial latent pivots. A two-stage autoregressive model first generates mesh vertices and then predicts triangular faces conditioned on the generated vertices and the same sparse latent pivots (Fig.~\ref{fig:teaser}).

Crucially, the same $16$ latent pivots provide geometric conditioning to both
generation stages, forming a compact shared structural interface between the
input point cloud and the mesh decoders. This reduces the geometric
conditioning set by $128\times$ relative to the $2{,}048$ input points and
uses a $16.1\times$ shorter decoder-conditioning sequence than recent
point-cloud-conditioned autoregressive methods using $257$ conditioning
tokens. The compact conditioning also reduces the amount of geometric context
consumed repeatedly during autoregressive decoding while preserving spatial
and learned geometric information.

Our main contributions are:

\begin{enumerate}

\item We propose \textbf{OptimusMesh}, a direct point-cloud-to-mesh framework
that autoregressively generates compact triangle meshes by predicting vertices
followed by triangular faces, without relying on implicit surface extraction
or costly post-hoc simplification.

\item We introduce a two-stage vertex-and-face generation architecture with a
\textbf{shared sparse geometric conditioning interface}, where the same latent
pivots guide vertex generation and subsequently face generation together with
the generated vertices.

\item We design a sparse latent pivot representation that compresses
$2{,}048$ oriented points into only $16$ pivots, reducing the geometric
conditioning set by $128\times$ and using a $16.1\times$ shorter
decoder-conditioning sequence than recent point-cloud-conditioned
autoregressive methods. Experiments show $25.7\%$--$94.1\%$ fewer faces than
the compared recent autoregressive methods while maintaining competitive
geometric fidelity and distributional quality.

\end{enumerate}

\begin{figure*}[t]
  \centering
  \includegraphics[width=\textwidth]{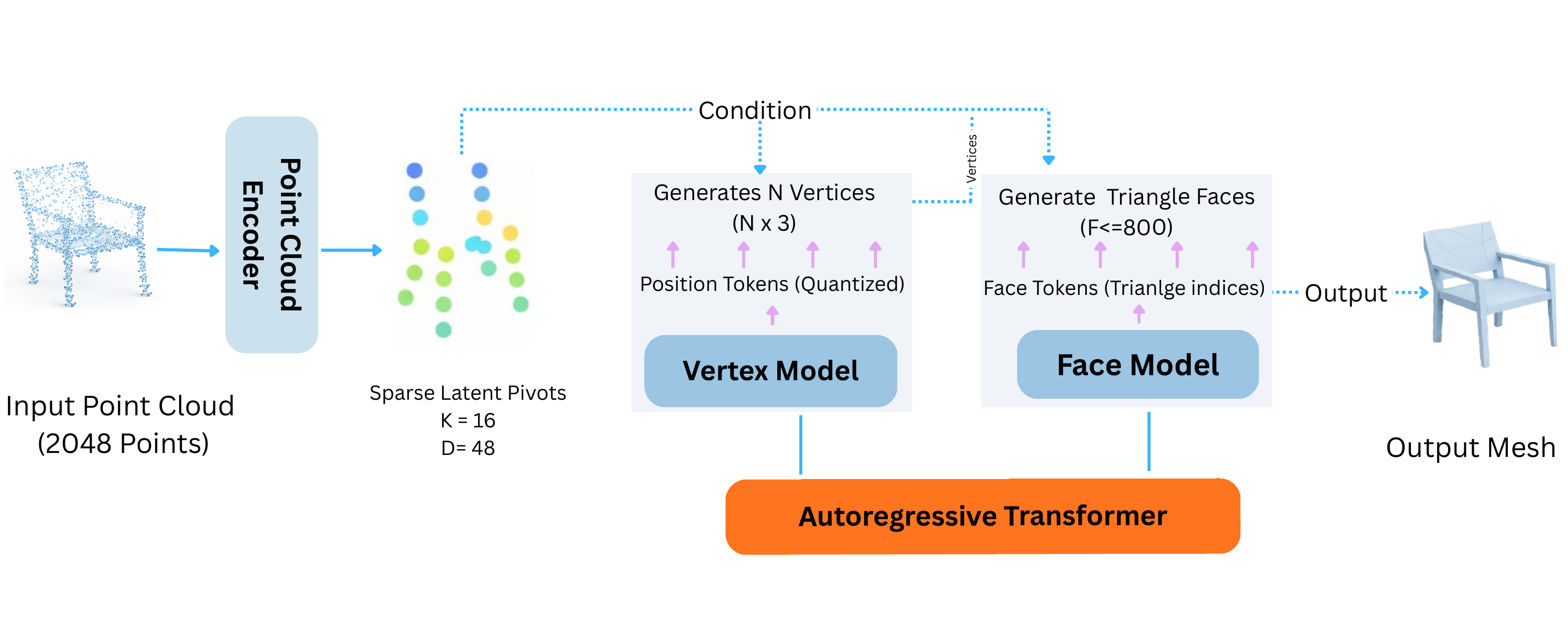}
  \caption{\textbf{OptimusMesh pipeline.}
  Given $2{,}048$ oriented input points, the point-cloud encoder extracts
  $K{=}16$ sparse latent pivots with $D{=}48$ features, reducing the geometric
  conditioning sequence by $128\times$. The vertex model first generates
  $V\in\mathbb{R}^{N\times3}$, followed by a face model conditioned on the
  generated vertices and the same pivots to produce a compact triangle mesh
  with $F\leq800$.}
  \label{fig:pipeline}
\end{figure*}
%% ============================================================
%% Related Work
%% ============================================================
\section{Related Work}
\label{sec:related}

\subsection{3D Shape Representations and Mesh Extraction}

3D shapes can be represented explicitly as meshes or point clouds, or
implicitly through continuous fields. Neural implicit
representations~\cite{Chen2019ImplicitFields,Mescheder2019OccupancyNetworks,
Park2019DeepSDF} have shown strong performance for object and scene
modeling~\cite{Jiang2020LocalImplicitGrid,Peng2020ConvONet,Sitzmann2020SIREN}.
Likewise, generative approaches operate on
voxels~\cite{Brock2016VoxelCNN,Wu2016Learning3DGAN},
point clouds~\cite{Luo2021DiffusionPointCloud,Zeng2022LION,
Zhou2021PointVoxelDiffusion}, or neural 3D
representations~\cite{Gao2022GET3D,Gupta2023TriplaneLatentDiffusion3DGen,
Li2023DiffusionSDF}. To obtain explicit meshes, these representations commonly
require surface extraction such as Marching
Cubes~\cite{Lorensen1987MarchingCubes} or related
methods~\cite{Shen2021DeepMarchingTetrahedra}. Recent pipelines based on
triplanes~\cite{Hong2024LRM,Li2024Instant3D},
FlexiCubes~\cite{Xu2024InstantMesh}, or intermediate point
representations~\cite{Lyu2023SLIDE} follow a similar paradigm. Although
effective, the resulting meshes can be dense or require additional processing
and simplification. Our work instead directly generates compact triangle meshes
from point-cloud inputs.

\subsection{Point Cloud to Mesh Reconstruction}

Classical methods such as the Ball Pivoting Algorithm
(BPA)~\cite{Bernardini1999BPA} and Screened Poisson Surface Reconstruction
(PSR)~\cite{Kazhdan2013ScreenedPoisson} recover surfaces directly from point
samples, but are sensitive to input density and sampling quality.
Learning-based methods such as Shape-as-Points
(SAP)~\cite{Peng2021ShapeAsPoints} predict an implicit indicator field
followed by surface extraction, typically producing high-resolution meshes.
Point2Mesh~\cite{Hanocka2020Point2Mesh} instead optimizes a mesh for each input
shape through iterative refinement, providing strong single-shape fitting at
the cost of per-shape optimization. In contrast, OptimusMesh directly decodes
compact meshes from point clouds without dense surface extraction or test-time
optimization.

\subsection{Autoregressive Mesh Generation}
Autoregressive mesh generation directly models explicit mesh structure without
relying on intermediate volumetric or implicit representations.
PolyGen~\cite{Nash2020PolyGen} factorizes mesh generation into autoregressive
vertex prediction followed by face generation, while
MeshGPT~\cite{Siddiqui2024MeshGPT} introduces learned discrete mesh tokens
through vector quantization. MeshAnything~\cite{Chen2025MeshAnything} and
MeshAnythingV2~\cite{Chen2025MeshAnythingV2} extend this direction to
point-cloud-conditioned artist meshes, with the latter introducing adjacent
mesh tokenization. PivotMesh~\cite{Weng2025PivotMesh} uses structured
pivot-vertex guidance, while EdgeRunner~\cite{Tang2025EdgeRunner} and
TreeMeshGPT~\cite{Lionar2025TreeMeshGPT} improve mesh serialization through
traversal- and tree-based sequencing.

Recent methods further address output-sequence complexity.
FastMesh~\cite{Kim2026FastMesh} decouples vertex and face generation to reduce
token redundancy and accelerate mesh generation, DeepMesh~\cite{Zhao2025DeepMesh} targets high-resolution meshes through improved tokenization and reinforcement-learning-based
post-training, and MeshRipple~\cite{Lin2026MeshRipple} introduces topology-aligned generation and long-range contextual modeling. In contrast, OptimusMesh focuses on the \emph{conditioning side}: it compresses $2{,}048$ oriented input points into only $K{=}16$ latent pivots, reducing the geometric conditioning set by $128\times$ and yielding a $16.1\times$ shorter decoder-conditioning sequence than recent point-cloud-conditioned
autoregressive methods using $257$ conditioning tokens. The same sparse conditioning is shared by both vertex and face generation under bounded mesh complexity.

%-------------------------------------------------------------------------

%%%%%%%%%%%%methodology

%% ============================================================
%% Methodology
%% ============================================================
%% ============================================================
%% Methodology
%% ============================================================
\section{Methodology}
\label{sec:method}

\subsection{Overview}

\begin{figure*}[t]
  \centering
  \includegraphics[
    width=\textwidth,
    height=45mm,
    keepaspectratio
  ]{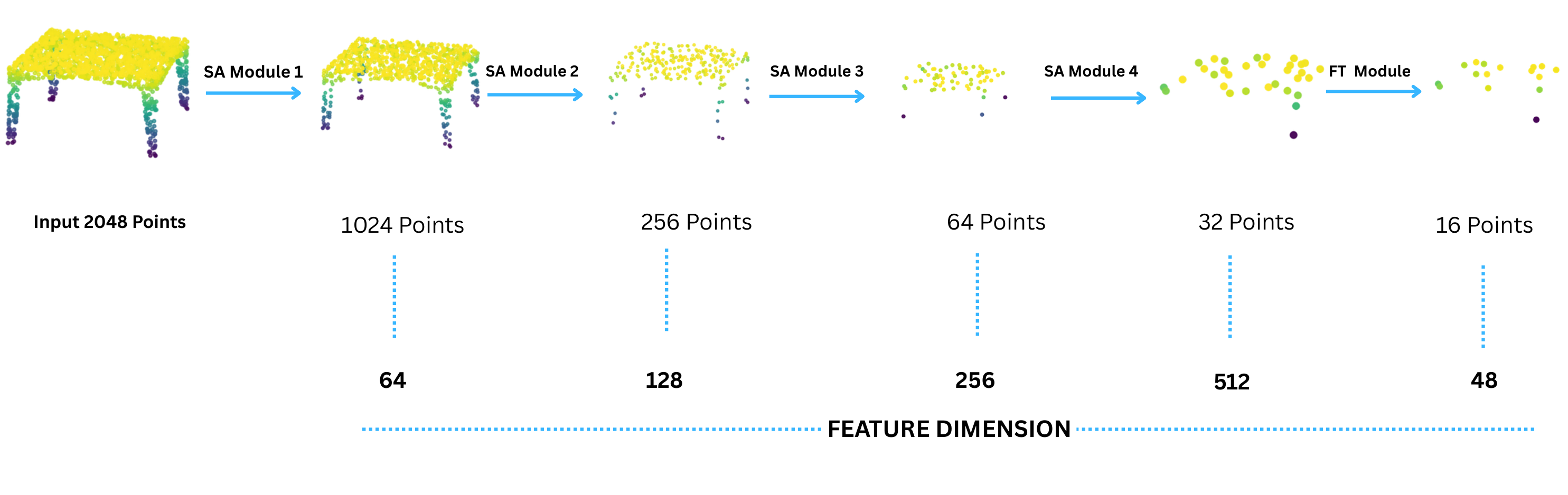}
  \caption{\textbf{Sparse latent pivot encoder.}
  The SLIDE-style point-cloud encoder progressively downsamples $2{,}048$
  oriented input points to $K{=}16$ sparse latent pivots with feature
  dimension $D{=}48$.}
  \label{fig:encoder}
\end{figure*}

We propose a framework for direct mesh generation from point clouds
(Fig.~\ref{fig:pipeline}). Given an input point cloud, we first extract a
sparse set of $K{=}16$ latent points, referred to as \emph{latent pivots},
using a point-cloud encoder (Fig.~\ref{fig:encoder}). Unlike dense geometric
conditioning, our framework compresses the input point cloud into a compact
latent interface shared across both decoding stages. These pivots condition
an autoregressive vertex model that generates mesh vertices, followed by a
face model that predicts triangular connectivity using both the generated
vertices and the same latent pivots.

Vertices are generated sequentially using a causal Transformer that predicts
a distribution over the next quantized coordinate token. The face model takes
the generated vertex set, latent pivots, and current sequence of face indices
as input and predicts vertex indices to form triangular faces. This two-stage
design directly produces compact meshes without relying on dense surface
extraction or post-processing.

\subsection{Sparse Latent Pivot Encoder}
\label{sec:pivot_encoder}
Our pipeline requires a point-cloud encoder that maps the input point cloud
into a sparse set of latent pivots, each associated with a feature embedding.
We adopt the sparse point-cloud encoder architecture of
SLIDE~\cite{Lyu2023SLIDE} and configure it to encode $2{,}048$ oriented input
points into $K{=}16$ latent pivots of dimension $D{=}48$. Each pivot contains
a 3D spatial coordinate and 45 learned feature channels. The encoder is trained from scratch on the mixed
ShapeNet~\cite{Chang2015ShapeNet},
Objaverse~\cite{Deitke2023Objaverse}, and
Toys4K~\cite{Stojanov2021Toys4K} training set without category conditioning,
and is subsequently frozen during vertex and face decoder training.
As illustrated in Fig.~\ref{fig:encoder}, the encoder consists of four Set
Abstraction (SA) modules~\cite{Lyu2022ConditionalPointDiffusion}, a lightweight
PointNet++-style branch~\cite{Lyu2022ConditionalPointDiffusion}, and a feature
transfer (FT) module that maps multi-scale features to the $K{=}16$ latent
pivots.
The input to each SA module is a set of points with per-point features. Each
module first subsamples the points using farthest point sampling (FPS).
Features are then propagated to the subsampled points by retrieving $k$
nearest neighbors, transforming their features with a shared MLP, and
aggregating them using an attention mechanism. The SA hierarchy progressively
downsamples the input point cloud while extracting multi-scale geometric
features. In our setup, each input point contains its 3D coordinates and
surface normal. We refer readers to
PDR~\cite{Lyu2022ConditionalPointDiffusion} for the full SA definition.

The encoder transfers multi-scale features from the final SA hierarchy to the
latent pivot locations using the FT module proposed in
PDR~\cite{Lyu2022ConditionalPointDiffusion}. Since the FT module requires
initial features at the target locations, a lightweight PointNet++-style
branch first extracts initial pivot features. Features from the final SA level
are then transferred to the pivot locations and concatenated with these
initial features to produce the final latent pivot embeddings. The resulting
representation captures both local geometric information and global shape
structure.

While SLIDE~\cite{Lyu2023SLIDE} uses this sparse representation for point-cloud
generation in latent space, we use the resulting pivots as compact geometric
conditioning for both autoregressive vertex and face decoding.

\subsection{Autoregressive Vertex Model}
\label{sec:vertex_model}

Given sparse latent pivots $Z=\{z_1,\dots,z_{16}\}$ extracted from the input
point cloud, the vertex model defines a conditional autoregressive distribution
over a flattened sequence of quantized coordinate tokens
$V^{\mathrm{seq}}=(v_1,\dots,v_T)$:
\begin{equation}
  p(V^{\mathrm{seq}} \mid Z;\theta)
  =
  \prod_{t=1}^{T}
  p\!\left(v_t \mid v_{<t}, Z;\theta\right).
  \label{eq:vertex_ar}
\end{equation}
where $T$ is the token-sequence length. The model predicts quantized coordinate
bins directly and terminates generation using an end-of-sequence token
\texttt{EOS}.

\paragraph{Coordinate tokenization.}
Each mesh is normalized to the canonical range $[-0.5,0.5]^3$ and processed
using a PolyGen-style quantization pipeline. Following
PolyGen~\cite{Nash2020PolyGen}, we apply 8-bit uniform quantization to each
vertex coordinate, resulting in $256$ bins per axis
~\cite{vanDenOord2016PixelRNN}. The quantized coordinates therefore lie in
$\{0,\dots,255\}$. Vertices are serialized as coordinate triplets in
\emph{$z,y,x$} order and followed by \texttt{EOS}:
\begin{equation}
  V^{\mathrm{seq}}
  =
  (z_1,y_1,x_1,\,
   z_2,y_2,x_2,\,
   \dots,\,
   z_N,y_N,x_N,\,
   \texttt{EOS}).
  \label{eq:vertex_seq}
\end{equation}
where $N$ is the number of vertices. \texttt{PAD} is used only for batching;
padded positions are ignored in the loss and are not valid prediction targets.
The prediction vocabulary therefore consists of $256$ coordinate bins plus
\texttt{EOS}, giving $257$ classes. 

\paragraph{Pivot-conditioned Transformer decoding.}
We model $V^{\mathrm{seq}}$ using a causal Transformer decoder. Each input
embedding is the sum of: (i) a token embedding, (ii) a vertex-position
embedding indicating which vertex the coordinate belongs to, shared across
its $z,y,x$ triplet, and (iii) a coordinate-slot embedding indicating whether
the token corresponds to $z$, $y$, or $x$. A learned \texttt{BOS} embedding
is prepended to start generation.

The decoder operates on quantized coordinate tokens, while the continuous
pivot features are projected into the decoder hidden space and supplied as
cross-attention memory. In addition, the mean pivot representation is
projected and added as a global bias to the decoder token embeddings.

\paragraph{Output distribution.}
Let $h_t$ denote the decoder hidden state at step $t$. The model predicts
logits over the $256$ coordinate bins and \texttt{EOS}:
\begin{equation}
  \ell_t = W h_t,\qquad
  p(v_t \mid v_{<t}, Z)=\mathrm{softmax}(\ell_t).
  \label{eq:vertex_out}
\end{equation}

\paragraph{Training.}
We train the vertex model with teacher forcing using cross-entropy over valid
(non-\texttt{PAD}) positions in the flattened coordinate-token sequence.
During training, token-level input dropout is applied to the decoder inputs.
To reduce exposure bias, a small fraction of coordinate tokens can also be
corrupted using nearby quantized perturbations or uniformly sampled coordinate
bins. The loss can additionally reweight $z$-coordinate and \texttt{EOS}
positions to improve axis balance and termination stability.

\paragraph{Inference and stopping.}
At inference time, tokens are generated autoregressively using temperature
$T_v{=}0.8$ and top-$p{=}0.95$ sampling until \texttt{EOS} is emitted.
To ensure valid termination, \texttt{EOS} is allowed only at complete vertex
boundaries, when the next token corresponds to a $z$ coordinate. Generation is
limited to at most $900$ vertices. The generated coordinates are then
dequantized back to the continuous normalized mesh space.

\subsection{Autoregressive Face Model}
\label{sec:face_model}

Given a vertex set $V=\{v_i\}_{i=1}^{|V|}$ and sparse latent pivots $Z$, the
face model defines a conditional autoregressive distribution over a serialized
face-token sequence $F^{\mathrm{seq}}=(f_1,\dots,f_T)$:
\begin{equation}
  p(F^{\mathrm{seq}} \mid V,Z;\theta)
  =
  \prod_{t=1}^{T}
  p\!\left(f_t \mid f_{<t},V,Z;\theta\right).
  \label{eq:face_ar}
\end{equation}
where $T$ denotes the face-token sequence length.

\paragraph{Triangle-only tokenization.}
Connectivity is represented strictly using triangles. Each triangle is
serialized as
$(\texttt{NEW\_TRI},v_a,v_b,v_c)$, and generation terminates with
\texttt{STOP}. We use the token set
\begin{equation}
\begin{aligned}
\mathcal{V}_{F}
&=
\{\texttt{STOP},\texttt{NEW\_TRI},\texttt{PAD}\} \\
&\quad \cup
\{\texttt{VERT\_OFFSET}+i
\mid i=0,\ldots,|V|-1\}.
\end{aligned}
\label{eq:face_vocab}
\end{equation}
where \texttt{PAD} is used only for batching. Padded positions are ignored
in the loss and are not valid prediction targets.

\paragraph{Contextual vertex token embeddings.}
Following the pointer-based autoregressive formulation of
PolyGen~\cite{Nash2020PolyGen}, we encode the vertices into a token table that
supports prediction over a variable-sized vertex set. Each vertex coordinate
is uniformly quantized to 8-bit bins over $[-0.5,0.5]$ and embedded using
separate per-axis embedding tables. The three coordinate embeddings are summed
to obtain a per-vertex representation.

Learned embeddings for \texttt{STOP} and \texttt{NEW\_TRI} are prepended to
the valid vertex sequence:
\begin{equation}
  [\texttt{STOP},\texttt{NEW\_TRI},v_1,\dots,v_{|V|}].
  \label{eq:vertex_token_table}
\end{equation}
This sequence is passed through a Transformer encoder, allowing self-attention
to contextualize each token with respect to the remaining vertices and special
tokens. The resulting outputs form contextual token embeddings $\{e_k\}$ for
\texttt{STOP}, \texttt{NEW\_TRI}, and each valid vertex index. A learned
\texttt{PAD} embedding is introduced only afterward for batched decoding and
is masked during both training and prediction.

\par\smallskip
\paragraph{Pointer-based face decoding.}
We use a causal Transformer decoder with cross-attention to the conditioning
memory. Each input face token is represented by gathering its contextual
embedding $\mathbf{e}_k$ from the token table and adding triangle-index and
within-triangle slot embeddings.

Let $p_t(k)=p(f_t=k\mid f_{<t},V,Z)$. Pointer logits and the corresponding
distribution are computed as
\begin{equation}
\begin{aligned}
\ell_{t,k}
&=
\frac{(W_p\mathbf{h}_t)^\top\mathbf{e}_k}{\sqrt{d}},\\
p_t(k)
&=
\frac{\exp(\ell_{t,k})}
{\sum_{j\in\mathcal{A}_t}\exp(\ell_{t,j})},
\qquad k\in\mathcal{A}_t .
\end{aligned}
\label{eq:face_distribution}
\end{equation}

During teacher-forced training, \texttt{PAD} and padded vertex entries are
masked, but no slot-specific grammar mask is applied. During inference,
slot-specific grammar masking permits only \texttt{NEW\_TRI} and
\texttt{STOP} at triangle boundaries and valid vertex pointers at
vertex-selection positions; repeated vertex indices within a triangle are
also masked.
\paragraph{Pivot-guided conditioning.}
The face model additionally conditions on the same $K{=}16$ sparse latent
pivots extracted from the input point cloud. Pivot features are projected to
the decoder hidden dimension and combined with the encoded vertex tokens to
form the decoder cross-attention memory. This provides global structural
guidance from the input point cloud together with local geometric information
from the generated vertices.

\paragraph{Training.}
We train the face model with teacher forcing by maximizing the likelihood of
the ground-truth face-token sequence while excluding padded positions from the
loss. During training, the model is conditioned on the ground-truth
preprocessed vertices and corresponding pivot latents. At inference time, it
instead receives vertices generated by the autoregressive vertex model together
with the same pivots and predicts connectivity autoregressively.

The \texttt{STOP} token can be upweighted to stabilize termination, while
vertex dropout can be applied during training to improve robustness to
imperfect generated vertices. We use a face-token budget of $3{,}201$ tokens,
corresponding to at most $800$ triangular faces followed by a final
\texttt{STOP}.

%% ============================================================
%% Experiments
%% ============================================================
\section{Experiments}
\label{sec:experiments}

\subsection{Experimental Setup}
\label{sec:exp_setup}

\noindent\textbf{Datasets and preprocessing.}

We train the mesh decoders on a mixed dataset of $35{,}907$ compact meshes
comprising $16{,}578$ ShapeNet~\cite{Chang2015ShapeNet},
$16{,}271$ Objaverse~\cite{Deitke2023Objaverse}, and
$3{,}058$ Toys4K~\cite{Stojanov2021Toys4K} samples. The combined dataset
contains $28{,}105$ training, $2{,}552$ validation, and $5{,}250$ test meshes.
We use ShapeNet as the primary quantitative benchmark for direct comparison
with prior mesh-generation and reconstruction methods.

Following prior autoregressive mesh modeling
~\cite{Nash2020PolyGen,Siddiqui2024MeshGPT,Weng2025PivotMesh}, we convert
meshes to triangular representations and construct compact targets suitable
for sequence modeling. For ShapeNet, we apply planar decimation in Blender
following prior autoregressive mesh preprocessing
~\cite{Nash2020PolyGen,Siddiqui2024MeshGPT}. For Objaverse, we follow the mesh
preprocessing practice of PivotMesh~\cite{Weng2025PivotMesh}, while retaining
naturally compact assets within our target complexity. For Toys4K, mesh
components are joined, triangulated, cleaned, and simplified using planar
decimation. Across all datasets, decoder targets are restricted to at most
$800$ triangular faces and $900$ vertices.

Meshes are normalized and quantized to 8 bits per coordinate before
autoregressive sequence construction. For geometric conditioning, we sample
$2{,}048$ oriented surface points per shape. Additional details on dataset construction, filtering, splits, and preprocessing are provided in the supplementary material.

\par\smallskip
\noindent\textbf{Evaluation metrics.}
We evaluate generated meshes using both distribution-level quality and paired
geometric fidelity. For distribution-level evaluation, we uniformly sample
$2{,}048$ points from each generated and reference mesh and report Minimum
Matching Distance based on Chamfer Distance (MMD-CD), coverage (COV-CD), and
the deviation of 1-NNA from its ideal value,
$\lvert1\text{-NNA}-0.5\rvert$. Lower MMD-CD indicates better distributional
fidelity, higher COV-CD indicates broader coverage, and a smaller 1-NNA
deviation indicates better agreement between generated and reference
distributions.

For paired geometric evaluation, each generated mesh is compared with its
corresponding ground-truth mesh. Both meshes are normalized, and $5{,}000$
points are uniformly sampled from each surface. We report symmetric Chamfer
Distance (CD-L1), Hausdorff Distance (HD), Normal Consistency (NC), and F1 at
$2\%$ of the ground-truth bounding-box diagonal. Mesh compactness is measured
by the average numbers of vertices (\#V) and faces (\#F).

For efficiency evaluation, we additionally report the number of input points, decoder conditioning tokens, and inference time. Conditioning tokens refer to the encoded geometric sequence supplied to the mesh decoder rather than the raw point-cloud size.

\par\smallskip
\noindent\textbf{Baselines.}
We compare OptimusMesh with recent point-cloud-conditioned autoregressive mesh
generators, including MeshAnything~\cite{Chen2025MeshAnything},
MeshAnythingV2~\cite{Chen2025MeshAnythingV2},
FastMesh~\cite{Kim2026FastMesh}, and
MeshRipple~\cite{Lin2026MeshRipple}. These methods form the primary comparison
for paired geometric fidelity. We additionally include Screened
PSR~\cite{Kazhdan2013ScreenedPoisson},
SAP~\cite{Peng2021ShapeAsPoints}, and
NKSR~\cite{Huang2023NKSR} in the distributional-quality and compactness
evaluation to provide reference against surface-reconstruction approaches.

\par\smallskip
\noindent\textbf{Implementation details.}
OptimusMesh consists of three components: (i) the sparse latent pivot encoder
(Sec.~\ref{sec:pivot_encoder}), (ii) the autoregressive vertex model
(Sec.~\ref{sec:vertex_model}), and (iii) the autoregressive face model
(Sec.~\ref{sec:face_model}). The point-cloud encoder is trained from scratch
on the mixed ShapeNet, Objaverse, and Toys4K training set without category
conditioning to produce $K{=}16$ latent pivots with feature dimension
$D{=}48$. Encoder training takes approximately 48 hours. The trained encoder is then kept frozen while training the mesh decoders. 

\paragraph{Vertex model.}
The vertex model uses a 12-layer Transformer decoder with hidden dimension
512, feed-forward dimension 2048, 8 attention heads, and dropout 0.1. It is
trained with teacher forcing on 8-bit quantized coordinate tokens using Adam
and a warmup-cosine learning-rate schedule. We train the vertex decoder with
a global batch size of 12 on $3\times$ NVIDIA RTX 6000 Ada GPUs for
approximately 45 hours.

\paragraph{Face model.}
The face model uses a 12-layer vertex encoder and a 12-layer masked Transformer
decoder with hidden dimension 256, feed-forward dimension 1024, and 8 attention
heads. It is trained with teacher forcing using AdamW and a progressive
face-budget curriculum up to $800$ triangles. We train the face decoder with
a global batch size of 8 on $4\times$ NVIDIA Tesla V100 32GB GPUs for
approximately 48 hours. After encoder training, the vertex and face decoders
are trained independently and can be trained in parallel.

\begin{figure*}[t]
  \centering
  \includegraphics[
    width=\textwidth,
    keepaspectratio
  ]{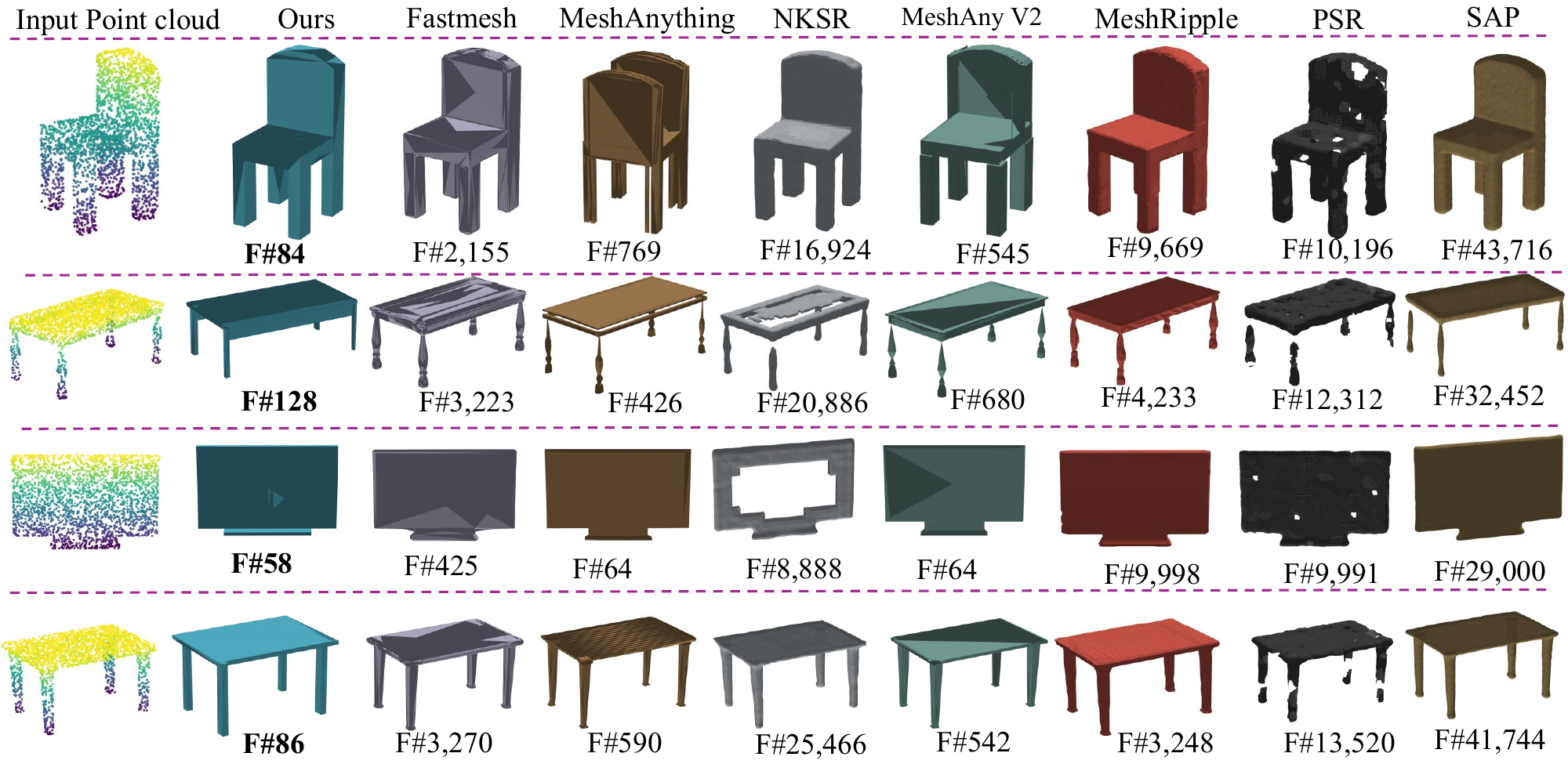}
  \caption{\textbf{Qualitative comparison on ShapeNet.}
  Representative point-cloud-conditioned mesh generation results from
  OptimusMesh and the evaluated baselines. Face counts (\#F) are reported
  below each generated mesh. OptimusMesh produces compact meshes while
  preserving the overall structure of the input shapes.}
  \label{fig:qualitative_shapenet}
\end{figure*}

\paragraph{Inference.}
At inference time, we use top-$p$ sampling for vertex and face generation.
Vertex generation terminates with \texttt{EOS}, while face generation follows
a bounded triangle budget of at most $800$ faces.

\subsection{Quantitative Results}
\label{sec:quantitative}

\Cref{tab:distribution_quality} reports distribution-level quality and mesh
compactness. OptimusMesh produces the most compact outputs, with only 141
vertices and 249 faces on average, and achieves the lowest 1-NNA deviation.
MeshAnything and FastMesh achieve the best MMD-CD and COV-CD, respectively.
OptimusMesh nevertheless outperforms the reconstruction-based methods in both
MMD-CD and COV-CD while using substantially fewer mesh elements.

\begin{table}[t]
  \caption{Distribution-level quality and mesh compactness on the ShapeNet test
subset. Best results are shown in \textbf{bold} and second-best results are
\underline{underlined}.}
  \label{tab:distribution_quality}
  \centering
  \small
  \setlength{\tabcolsep}{2.2pt}
  \begin{tabular}{@{}lccccc@{}}
    \toprule
    Method &
    MMD $\downarrow$ &
    COV $\uparrow$ &
    1-NNA $\downarrow$ &
    \#V $\downarrow$ &
    \#F $\downarrow$ \\
    \midrule

    PSR
      & 0.0849 & 0.4989 & 0.1203 & 5,980 & 11,563 \\

    SAP
      & 0.0863 & 0.4855 & 0.1392 & 20,957 & 41,912 \\

    NKSR
      & 0.0831 & 0.4922 & 0.1025 & 8,904 & 17,694 \\

    \midrule

    MeshAnything
      & \textbf{0.0659}
      & \underline{0.6347}
      & 0.1492
      & \underline{182}
      & \underline{335} \\

    MeshAnythingV2
      & 0.0689
      & 0.6125
      & 0.1192
      & 463
      & 858 \\

    FastMesh-V1K
      & \underline{0.0660}
      & \textbf{0.6659}
      & 0.1481
      & 609
      & 3,888 \\

    MeshRipple-10K
      & 0.0731
      & 0.5768
      & \underline{0.0869}
      & 2,132
      & 4,250 \\

    \midrule

    \textbf{Ours}
      & 0.0813
      & 0.5657
      & \textbf{0.0846}
      & \textbf{141}
      & \textbf{249} \\

    \bottomrule
  \end{tabular}
\end{table}

On the 450 common ShapeNet test objects (\Cref{tab:paired_geometry}), OptimusMesh achieves the lowest CD-L1 and HD and the highest NC among the compared recent mesh generators. MeshRipple
achieves the highest F1@2\%, while OptimusMesh maintains stronger overall surface-distance and normal agreement with substantially lower output complexity.
\begin{table}[t]
  \caption{Paired geometric fidelity on 450 common ShapeNet test objects. Best
results are shown in \textbf{bold} and second-best results are
\underline{underlined}.}
  \label{tab:paired_geometry}
  \centering
  \small
  \setlength{\tabcolsep}{3.2pt}
  \begin{tabular}{@{}lcccc@{}}
    \toprule
    Method &
    CD-L1 $\downarrow$ &
    HD $\downarrow$ &
    F1@2\% $\uparrow$ &
    NC $\uparrow$ \\
    \midrule

    MeshAnything
      & 0.2123
      & 0.3617
      & 0.3881
      & 0.5233 \\

    MeshAnythingV2
      & 0.2059
      & 0.3546
      & 0.3985
      & 0.5367 \\

    FastMesh-V1K
      & \underline{0.1878}
      & \underline{0.3323}
      & \underline{0.4151}
      & \underline{0.5690} \\

    MeshRipple-10K
      & 0.1994
      & 0.3448
      & \textbf{0.4245}
      & 0.5608 \\

    \midrule

    \textbf{Ours}
      & \textbf{0.1421}
      & \textbf{0.2705}
      & 0.3570
      & \textbf{0.5956} \\

    \bottomrule
  \end{tabular}
\end{table}

\Cref{tab:decoder_efficiency} compares geometric conditioning and inference
efficiency. OptimusMesh uses only 16 decoder-conditioning tokens compared with
257 for the evaluated released implementations, corresponding to a
$16.1\times$ shorter conditioning sequence. FastMesh achieves the lowest
inference time, while OptimusMesh is the second fastest.
methods.
\begin{table}[t]
 \caption{Conditioning and inference efficiency of recent
point-cloud-conditioned autoregressive mesh generators. Best results are shown
in \textbf{bold} and second-best results are \underline{underlined}.}
  \label{tab:decoder_efficiency}
  \centering
  \small
  \setlength{\tabcolsep}{4pt}
  \begin{tabular}{@{}lrrr@{}}
    \toprule
    Method &
    Input pts. &
    Cond. tokens $\downarrow$ &
    Time (s) $\downarrow$ \\
    \midrule

    MeshAnything
      & 4,096
      & 257
      & 35.61 \\

    MeshAnythingV2
      & 8,192
      & 257
      & 63.40 \\

    FastMesh-V1K
      & 8,192
      & 257
      & \textbf{8.58} \\

    MeshRipple-10K
      & 16,384
      & 257
      & 379.50 \\

    \midrule

    \textbf{Ours}
      & 2,048
      & \textbf{16}
      & \underline{15.75} \\

    \bottomrule
  \end{tabular}
\end{table}

Overall, OptimusMesh achieves the most compact outputs and the shortest
conditioning sequence, while preserving competitive geometric fidelity and
achieving the best CD-L1, HD, and NC among the compared mesh generators.

\begin{figure}[t]
  \centering
  \includegraphics[
    width=\columnwidth,
    keepaspectratio
  ]{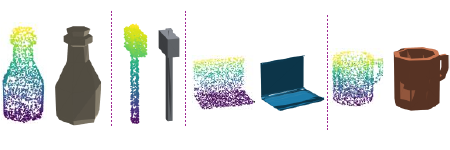}
  \caption{\textbf{Qualitative results on Toys4K.}
  Representative input point clouds and meshes generated by OptimusMesh
  across objects with diverse geometry and aspect ratios.}
  \label{fig:qualitative_toys4k}
\end{figure}

\subsection{Qualitative Results}
\label{sec:qualitative}

\Cref{fig:qualitative_shapenet} shows representative ShapeNet examples across
all evaluated methods. Surface-reconstruction approaches such as PSR, SAP,
and NKSR generally produce substantially denser outputs, while recent
autoregressive methods generate more compact meshes but still vary considerably
in output complexity. In contrast, OptimusMesh preserves the overall object
structure using only 58--128 faces in the examples shown. The comparison
illustrates the main design goal of our method: directly generating
low-complexity triangle meshes from sparse geometric conditioning while
retaining the dominant shape structure.

\Cref{fig:qualitative_toys4k} further shows representative results on Toys4K.
Despite the variation in object geometry and aspect ratio, OptimusMesh
reconstructs the overall shape from the input point clouds with compact
triangle meshes. These examples further illustrate that the learned sparse
pivot representation is applicable beyond the ShapeNet objects used in the
main qualitative comparison.

\subsection{Ablation Study}
\label{sec:ablation_study}

We ablate the number and density of geometric conditioning tokens in
\Cref{tab:sparse_dense_conditioning_ablation}. We compare $K{=}16$ and
$K{=}32$ sparse pivots with $2{,}048$ dense point features while keeping the
decoder architecture, training split, feature dimension ($D{=}48$), and
face budget fixed. The encoder is retrained accordingly for the $K{=}32$
setting. Relative to dense $2048{\times}48$ conditioning, the $16{\times}48$
pivot representation reduces the decoder-conditioning sequence by
$128\times$.

The $16$-pivot configuration performs best across all reported metrics while
using the shortest conditioning sequence. Increasing to $32$ pivots does not
improve reconstruction quality, whereas dense conditioning increases the
sequence length from $16$ to $2{,}048$ tokens and nearly doubles inference
time. These results support $K{=}16$ as a compact and effective conditioning
interface for both vertex and face generation.

\begin{table}[t]
  \caption{Ablation of sparse and dense geometric conditioning.}
  \label{tab:sparse_dense_conditioning_ablation}
  \centering
  \small
  \setlength{\tabcolsep}{2.0pt}
  \resizebox{\columnwidth}{!}{
  \begin{tabular}{@{}lrrrrrr@{}}
    \toprule
    Condition &
    Tok. $\downarrow$ &
    CD-L1 $\downarrow$ &
    F1@2\% $\uparrow$ &
    NC $\uparrow$ &
    \#F $\downarrow$ &
    Time $\downarrow$ \\
    \midrule

    $32{\times}48$ pivots
      & 32
      & 0.1789
      & 0.3012
      & 0.5423
      & 307.3
      & 28.10 \\

    Dense $2048{\times}48$
      & 2,048
      & 0.1647
      & 0.3026
      & 0.5584
      & 317.2
      & 34.66 \\

    \midrule

    \textbf{$16{\times}48$ pivots}
      & \textbf{16}
      & \textbf{0.1417}
      & \textbf{0.3571}
      & \textbf{0.5741}
      & \textbf{249.5}
      & \textbf{15.81} \\

    \bottomrule
  \end{tabular}}
\end{table}

\section{Conclusion}
\label{sec:conclusion}

We presented \textbf{OptimusMesh}, a direct point-cloud-to-mesh framework
using sparse latent pivot conditioning and autoregressive vertex and face
generation. By compressing $2{,}048$ oriented input points into only
$K{=}16$ latent pivots, our method reduces geometric conditioning while
directly generating compact triangle meshes with bounded complexity.
Experiments show that OptimusMesh produces the most compact outputs and the
shortest conditioning sequence among the compared methods while maintaining
competitive geometric fidelity, demonstrating the effectiveness of sparse
geometric conditioning for compact mesh generation.

%%%%%%%%% REFERENCES
{
    \small
    \bibliographystyle{ieeenat_fullname}
    \bibliography{main}
}

%%%%%%%%% SUPPLEMENTARY MATERIAL
\clearpage
\setcounter{page}{1}
\maketitlesupplementary

\section{Additional Qualitative Results}
\label{sec:supp_qualitative}

\begin{figure*}[t]
  \centering
  \includegraphics[width=\textwidth]{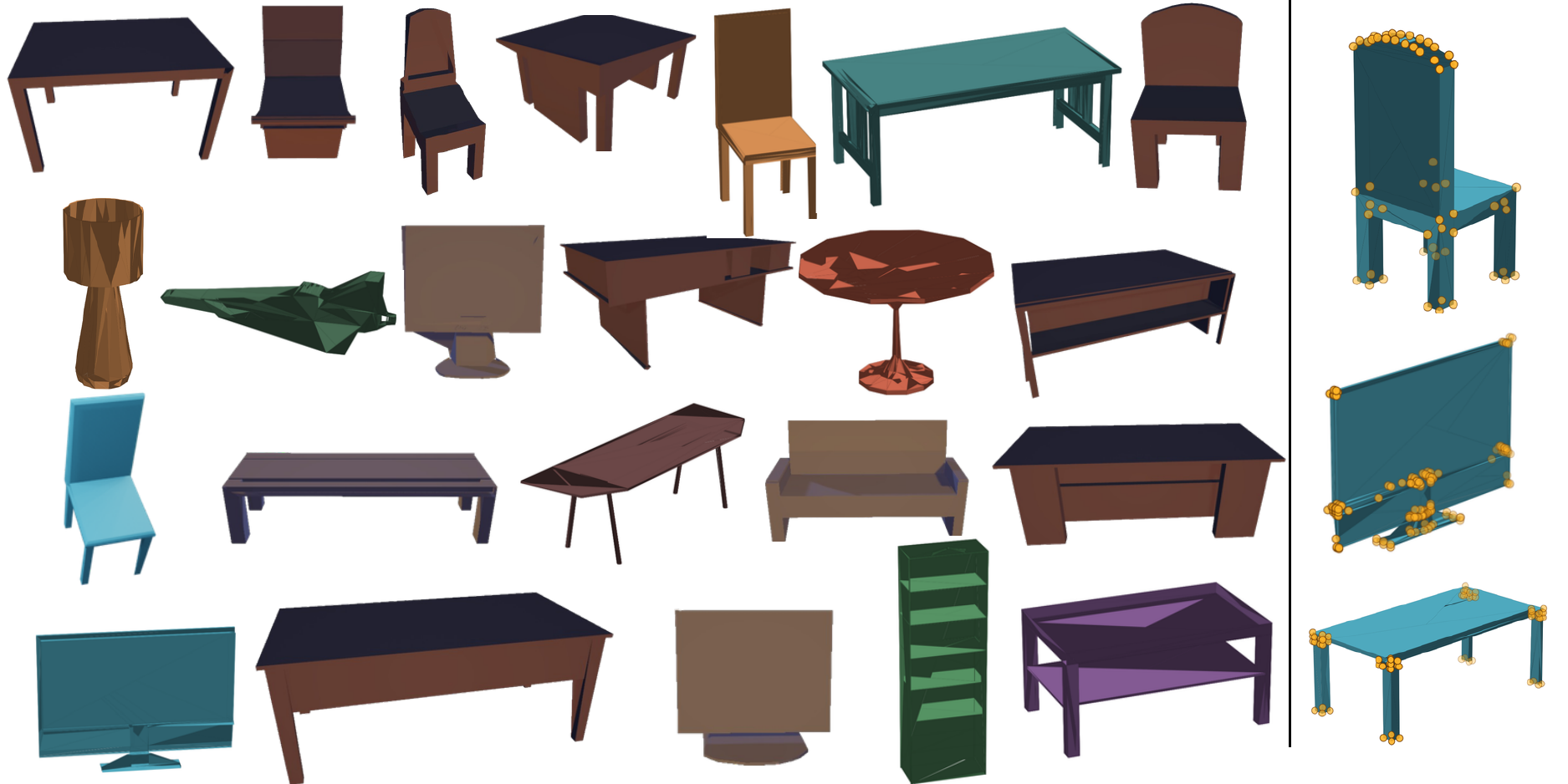}
  \caption{\textbf{Additional meshes generated by OptimusMesh.}
  Left: example output meshes generated by our method. Right: corresponding
  generated vertex sets before face prediction, illustrating how the predicted
  vertices are converted into final triangle meshes.}
  \label{fig:supp_additional_results}
\end{figure*}

Figure~\ref{fig:supp_additional_results} shows additional OptimusMesh outputs
together with representative generated vertex sets before face prediction.

\section{Dataset and Preprocessing Details}
\label{sec:supp_dataset}

\subsection{Dataset Composition}

We construct our data from ShapeNet~\cite{Chang2015ShapeNet},
Objaverse~\cite{Deitke2023Objaverse}, and
Toys4K~\cite{Stojanov2021Toys4K}. Table~\ref{tab:supp_dataset_composition}
summarizes the point-cloud corpus used for sparse-pivot encoder training and
the compact-mesh corpus used for vertex and face decoder training.

The encoder corpus is larger because encoder training requires only a valid
oriented point cloud, whereas decoder training additionally requires a compact
triangular target satisfying our complexity constraints. Source-level
train/validation/test assignments are preserved throughout preprocessing.

\subsection{ShapeNet}

We use 13 categories: airplane, bench, cabinet, car, chair, display, lamp,
speaker, rifle, sofa, table, telephone, and vessel. Oriented point clouds
contain surface positions and normals.

For decoder targets, we follow prior autoregressive mesh preprocessing
~\cite{Nash2020PolyGen,Siddiqui2024MeshGPT}. Meshes are triangulated and
simplified using planar decimation in Blender~\cite{Blender2018}. When
multiple valid compact variants are available, one is randomly selected during
training, while validation and test selection are deterministic. Targets
contain at most $800$ faces and $900$ vertices.

\subsection{Objaverse}

For Objaverse, we retain naturally compact GLB assets after applying
scene-graph transformations and welding geometrically duplicated vertices.
Only valid triangular meshes with $50$--$800$ faces and at most $900$
vertices are retained; no simplification is applied.

Meshes are centered and scaled such that the longest bounding-box dimension
equals one, after which surface points and normals are sampled. Objects are
assigned deterministically to approximately $85\%/5\%/10\%$
training/validation/test partitions.

\subsection{Toys4K}

For Toys4K, mesh components are joined, triangulated, cleaned, and normalized.
Surface points and normals are sampled from the normalized source mesh before
simplification.

Compact targets are produced using planar or edge-collapse simplification and
restricted to $20$--$800$ faces and at most $900$ vertices. Candidate meshes
are selected using bidirectional nearest-surface error, with 95th-percentile
error at most $0.035$ and maximum error at most $0.12$. Objects are split
deterministically into approximately $85\%/5\%/10\%$
training/validation/test partitions.
\begin{table*}[t]
  \caption{Dataset composition for sparse-pivot encoder and mesh-decoder
  training.}
  \label{tab:supp_dataset_composition}
  \centering
  \small
  \setlength{\tabcolsep}{4.0pt}
  \begin{tabular}{@{}lrrrrrrrr@{}}
    \toprule
    & \multicolumn{4}{c}{Point-cloud encoder}
    & \multicolumn{4}{c}{Mesh decoders} \\
    \cmidrule(lr){2-5}\cmidrule(lr){6-9}
    Dataset &
    Train & Val. & Test & Total &
    Train & Val. & Test & Total \\
    \midrule
    ShapeNet
    & 30,465 & 4,342 & 8,710 & 43,517
    & 11,675 & 1,617 & 3,286 & 16,578 \\
    Objaverse
    & 13,834 & 782 & 1,655 & 16,271
    & 13,834 & 782 & 1,655 & 16,271 \\
    Toys4K
    & 2,596 & 153 & 309 & 3,058
    & 2,596 & 153 & 309 & 3,058 \\
    \midrule
    \textbf{Total}
    & \textbf{46,895} & \textbf{5,277} & \textbf{10,674}
    & \textbf{62,846}
    & \textbf{28,105} & \textbf{2,552} & \textbf{5,250}
    & \textbf{35,907} \\
    \bottomrule
  \end{tabular}
\end{table*}

\subsection{Point-Cloud and Decoder Processing}

Across all datasets, $2{,}048$ oriented surface points are sampled per object.
The SLIDE-style encoder~\cite{Lyu2023SLIDE} is trained from scratch on the
mixed training set without category conditioning and produces $K{=}16$ latent
pivots, each containing a 3D position and 45 learned feature channels. The
trained encoder is frozen during decoder training.

Decoder meshes are mean-centered and scaled to approximately
$[-0.5,0.5]^3$, followed by 8-bit coordinate quantization. Duplicate and
unreferenced vertices are removed, vertices are ordered by $(z,y,x)$, and face
indices are remapped accordingly. The vertex model supports at most $900$
vertices ($2{,}701$ tokens including \texttt{EOS}); after quantization and
filtering, $33{,}822$ meshes remain for vertex training. The face model uses
the same processed vertices and supports at most $800$ triangles
($3{,}201$ tokens including \texttt{STOP}).

\section{Data Augmentation and Regularization}
\label{sec:supp_augmentation}

All stochastic augmentations are applied only during training.

\subsection{Sparse-Pivot Encoder}

Encoder training uses random anisotropic scaling up to $1.2$, reflection with
probability $0.5$, and translation with magnitude $0.1$. Farthest-point
sampling initializes the $16$ pivot locations. The training objective is

\begin{equation}
\mathcal{L}_{\mathrm{AE}}
=
\mathcal{L}_{\mathrm{CD}}
+
0.1\,\mathcal{L}_{\mathrm{feature}}
+
10^{-5}\mathcal{L}_{\mathrm{KL}}.
\end{equation}

\subsection{Vertex Decoder}

When multiple compact variants are available, one is randomly selected during
training. We apply global translation within $[-0.02,0.02]$ per axis and
token dropout with probability $0.1$. We additionally corrupt $10\%$ of
teacher-forcing coordinate tokens: $80\%$ using nearby quantized values with
Gaussian noise and $20\%$ using random coordinate symbols. Mixed-data sampling follows a ShapeNet/Objaverse/Toys4K ratio of $50/40/10$. We use cross-entropy with label smoothing $0.1$, Transformer dropout $0.1$,
and gradient clipping at norm $1.0$. Adam with linear warm-up and cosine decay
is used, with learning rates $3\times10^{-4}$ initially and
$2\times10^{-5}$ for mixed-data fine-tuning.

\subsection{Face Decoder}

Face training uses random compact-mesh selection and source-balanced sampling
with weights proportional to $n_s^{-0.5}$. Valid vertex-conditioning entries
are dropped with probability $0.02$, Transformer dropout is $0.2$, and
canonical vertex and face ordering is used without random vertex permutation.

The model is optimized using AdamW with weight decay $0.01$, linear warm-up
followed by cosine decay, and gradient clipping at norm $1.0$. Validation and
test processing are deterministic.

\section{Inference Sampling and Decoding}
\label{sec:supp_sampling}

\noindent\textbf{Vertex decoding.}
We use temperature $T_v{=}0.8$ and top-$p{=}0.95$ sampling. \texttt{EOS} is
allowed only at complete vertex boundaries, with generation limited to $900$
vertices. Before face decoding, generated vertices are clipped to
$[-0.5,0.5]$, quantized, deduplicated, ordered by $(z,y,x)$, and dequantized.

\smallskip
\noindent\textbf{Face decoding.}
The face decoder uses temperature $T_f{=}0.55$ and top-$p{=}0.85$. At triangle
boundaries, only \texttt{NEW\_TRI} and \texttt{STOP} are permitted; at
vertex-selection positions, only valid vertex pointers are permitted, with
repeated indices masked. \texttt{STOP} is enabled after at least 20 triangles,
and generation is capped at $800$ triangles.

The \texttt{STOP} logit uses a ramped bias of $0.18$ with power $2.0$.
Edge-consistency and triangle-area biases of $0.015$ and $0.006$ are applied
during sampling. A single frozen decoding configuration is used for all
reported experiments without per-object parameter or seed selection.

\smallskip
\noindent\textbf{Mesh export.}
Generated tokens are converted directly into indexed triangles. Degenerate
and duplicate faces, invalid indices, non-finite values, and unreferenced
vertices are removed. No mesh simplification, surface reconstruction,
test-time optimization, or ground-truth-based alignment is applied.

\section{Analysis of Sparse Latent Pivots}
\label{sec:supp_pivot_analysis}

Each $2{,}048$-point input is compressed into $16$ sparse latent pivots, each
containing a 3D position and 45 learned feature channels. The same pivots
condition both decoders, reducing the geometric conditioning set by
$128\times$.

\begin{figure}[t]
  \centering
  \includegraphics[width=\columnwidth]{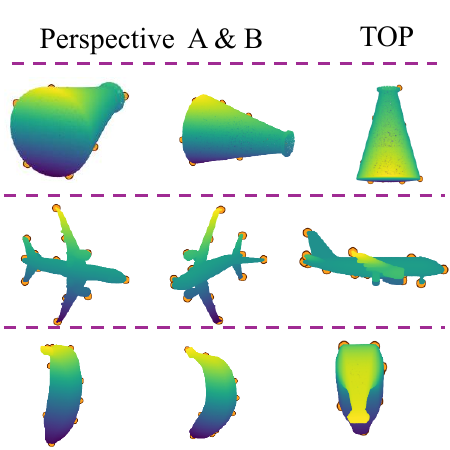}
  \caption{\textbf{Visualization of sparse latent pivots.}
  Representative objects are shown from multiple viewpoints. Orange markers
  denote the extracted pivot locations distributed across the input geometry.}
  \label{fig:supp_sparse_pivots}
\end{figure}

Figure~\ref{fig:supp_sparse_pivots} shows the extracted pivot locations from
multiple viewpoints. The pivots remain distributed across structurally
distinct regions rather than collapsing to a small area. Their positions
should not be interpreted as a reconstructed point cloud, since each pivot
also carries learned geometric features aggregated from the input.

\section{Decoding Failure Modes}
\label{sec:supp_failure}

\begin{figure}[t]
  \centering
  \includegraphics[width=\columnwidth]{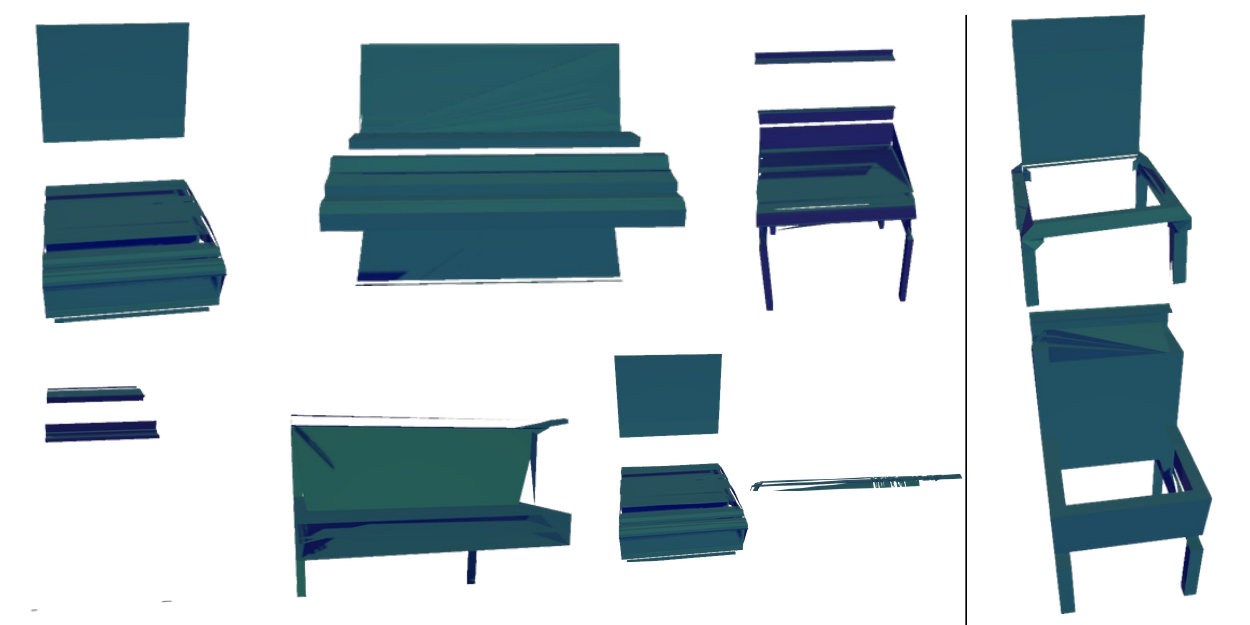}
  \caption{\textbf{Representative failure cases of OptimusMesh.}
  Examples include early termination, incomplete connectivity, and distorted
  local geometry.}
  \label{fig:supp_failure}
\end{figure}

Figure~\ref{fig:supp_failure} shows representative failure cases.
Autoregressive decoding can terminate early, producing incomplete geometry,
or late, producing redundant triangles. Thin structures may also be affected
by 8-bit quantization and vertex deduplication. All examples use the same
frozen decoding configuration as the quantitative evaluation.

\end{document}